\documentclass[conference]{IEEEtran}
\usepackage{cite}
\usepackage{amsmath,amssymb,amsfonts}
\usepackage{algorithmic}
\usepackage{graphicx}
\usepackage{textcomp}
\usepackage{xcolor}
\usepackage{placeins}
\usepackage{hyperref}
\usepackage{url}
\usepackage{svg}
\usepackage{calc}
\usepackage{mathptmx} 
\usepackage{comment}
\usepackage{fancyhdr}
\usepackage[normalem]{ulem}

\usepackage[final]{microtype}
\usepackage{flushend}
\usepackage{graphicx}
\usepackage[]{subfig}
\usepackage{xspace}
\usepackage{array}
\usepackage{xcolor}
\usepackage{makecell}
\usepackage{enumitem}
\usepackage{latexsym}
\usepackage{pifont}
\usepackage{array,booktabs}

\usepackage{booktabs} 
\usepackage[utf8]{inputenc} 
\usepackage[T1]{fontenc}    
\usepackage{booktabs}       
\usepackage{amsfonts}       
\usepackage{nicefrac}       
\usepackage{microtype}      
\usepackage{floatrow}

\definecolor{fabulous}{rgb}{1.0, 0.0, 0.5}

\newcommand{\half}{.5}

\newcommand{\third}{.33}

\newcommand\blfootnote[1]{%
  \begingroup
  \renewcommand\thefootnote{}\footnote{#1}%
  \addtocounter{footnote}{-1}%
  \endgroup
}

\usepackage{enumitem}
\setlist{nosep}
\def\BibTeX{{\rm B\kern-.05em{\sc i\kern-.025em b}\kern-.08em
    T\kern-.1667em\lower.7ex\hbox{E}\kern-.125emX}}

\begin{document}

\title{BitPruning: Learning Bitlengths for Aggressive and Accurate Quantization}

\author{\noindent
\resizebox{\linewidth}{!}{
\begin{tabular}{cccc}
\begin{tabular}[t]{c}
Milo\v{s} Nikoli\'{c}\\
\textit{University of Toronto}\\
Toronto, Canada
\end{tabular} &
\begin{tabular}[t]{c}
Ghouthi Boukli Hacene\\
\textit{Mila}\\
Montreal, Canada
\end{tabular} &
\begin{tabular}[t]{c}
Ciaran Bannon\\
\textit{University of Toronto}\\
Toronto, Canada
\end{tabular} &
\begin{tabular}[t]{c}
Alberto Delmas Lascorz\\
\textit{University of Toronto}\\
Toronto, Canada
\end{tabular} \\ \addlinespace[4ex] 
\begin{tabular}[t]{c}
Matthieu Courbariaux\\
\textit{Mila}\\
Montreal, Canada
\end{tabular} &
\begin{tabular}[t]{c}
Yoshua Bengio\\
\textit{Mila}\\
Montreal, Canada\end{tabular} &
\begin{tabular}[t]{c}
Vincent Gripon\\
\textit{IMT-Atlantique}\\
Brest, France
\end{tabular} &
\begin{tabular}[t]{c}
Andreas Moshovos\\
\textit{University of Toronto}\\
Toronto, Canada
\end{tabular}
\end{tabular}
}}

\maketitle

\begin{abstract}
Neural networks have demonstrably achieved state-of-the art accuracy using low-bitlength \mbox{integer} quantization, yielding both execution time and energy benefits on existing hardware designs that support short bitlengths. However, the question of finding the minimum bitlength for a desired accuracy remains open. We introduce a training method for minimizing inference bitlength at any granularity while maintaining accuracy. Namely, we propose a regularizer that \mbox{penalizes} large bitlength representations throughout the architecture and show how it can be modified to minimize other quantifiable criteria, such as number of operations or memory footprint. We demonstrate that our method learns thrifty representations while maintaining accuracy. With ImageNet, the method produces an average per layer bitlength of 4.13, 3.76 and 4.36 bits on AlexNet, ResNet18 and MobileNet V2 respectively, remaining within 2.0\%, 0.5\% and 0.5\% of the base TOP-1 accuracy. 
\end{abstract}

\begin{IEEEkeywords}
Heterogeneous quantization, learned datatype, neural networks
\blfootnote{Correspondence to: Milo\v{s} Nikoli\'{c} $<$milos.nikolic@mail.utoronto.ca$>$.}
\end{IEEEkeywords}

\section{Introduction}
\label{sct:introduction}
Over the past two decades energy and power (energy over time) have emerged as the primary constraints for computing devices~\cite{dark-silicon} dictating execution time performance, operating costs, or up-time for virtually all device segments, from data centers to the edge. Accordingly, reducing Deep Neural Network (DNN) energy needs can yield a multitude of benefits such as improved latency for existing models, enable the deployment of more powerful models, or boost throughput and reduce operating costs for data centers. A key choice that dictates DNN energy consumption and thus execution time during inference is the choice of \textit{datatype}, that is the number of bits used per value (activations or weights) and their numerical interpretation (e.g., floating-point or fixed-point). For example, using 8b fixed-point vs. 32b floating-point reduces the energy of a multiply-accumulate by more than $23\times$~\cite{Horowitz:Energy}, whereas using 4b fixed-point more than halves energy compared to 8b and doubles compute bandwidth even on existing hardware~\cite{NVIDIA_TO}. Choosing a datatype may at first appear a simple task. However, doing so naively can leave a lot of the potential benefits untapped, or worse can yield a network that fails to converge.

While 32-bit floating point is definitely sufficient, many DNNs can use other lower cost floating- or fixed-point representations without sacrificing accuracy. For this reason, and while earlier graphics processors (GPUs) and general-purpose processors (CPUs) supported only a few datatypes, newer generations of GPUs and CPUs have been extended to support additional more energy efficient datatypes specifically targeting DNNs~\cite{NVIDIA_TO,INTEL_8b_WP,NVIDIA_Presentation}.
Support for multiple datatypes is a key feature of specialized hardware accelerators~\cite{TPUISCA17,Stripes-MICRO,Bit-Fusion,pragmatic}.

Energy efficiency improves when using a narrower datatype (e.g., 16b instead of 32b floating-point) or a simpler and less expensive to implement one (e.g., fixed-point vs. floating-point). More energy efficient datatypes yield benefits in two ways (see Table~\ref{tbl:Accelerators} for example accelerators):\\
\noindent{\textbf{Memory:}} Using narrower datatypes reduces the number of bits that need to be stored and transferred.  This greatly reduces energy as off-chip memory accesses are today one order of magnitude slower and two orders of magnitude more energy consuming compared to on-chip operations (accesses or computation). Doing so enables us to run larger and more powerful networks, to deploy devices with smaller on- and off-chip memories, and to enjoy higher performance especially when memory bandwidth is limited, as it is often the case today. Besides being able to directly benefit CPUs and GPUs today, this observation also motivated several low-cost memory compression techniques that directly take advantage of narrower datatypes at layer~\cite{Judd:Proteus:ICS}~\cite{Stripes-MICRO}~\cite{Bit-Fusion} or finer granularity~\cite{ShapeshifterMICRO}~\cite{Park2018EnergyEfficientNN}.
\\
\noindent{\textbf{Computation:}} The operations per cycle throughput of many hardware platforms scales inversely proportionally with datatype bit length. Most commodity general purpose or graphics processors (CPUs or GPUs)  support several bit lengths, e.g., 16b at 1x  and 8b at 2x throughput, whereas, spatially composable and bit-serial accelerators \cite{DPRed,Stripes-MICRO,LOOM,Proteus,Bit-Fusion,bilaniuk2019bit,BISMO} can support a full range of bit lengths. Moreover, using simpler datatypes such as fixed-point vs. floating-point can also amplify compute bandwidth. For example, many commodity CPUs and GPUs implement more fixed-point units than floating-point ones. In accelerators, using simpler arithmetic units allows us to pack more of them for the same area and for the same energy budget. For example, we can fit 32 8b fixed-point MAC units is in the same area as 5 floating-point ones.

Multiple techniques of determining optimal datatypes have been proposed due to the impact the choice has even on existing hardware. However, today we still usually choose a target datatype prior to training a network without help on how to navigate their complex relationship with accuracy, energy and execution time performance. Once we have a datatype --- typically a fixed-point representation of a desired bitlentgh --- we can use one of the many methods which can maximize the accuracy with a fixed bitlength~\cite{binaryconnect,PACT,TQT,LSSQ}. This approach deprives us of the benefits of tighter datatype selection or worse forces us to search the datatype selection space via trial-and-error at the great cost of training the model anew multiple times. 

Profiling-based techniques can find per-layer bitlengths delivering further benefits~\cite{judd:reduced,milosispass19}. However, beyond a certain bitlength choice, such post-training methods will start incurring accuracy loss. Further reduction in bitlengths is possible if fine-tuning can recover some accuracy losses. Similarly, some hand crafted quantization techniques can treat value groups differently using advance knowledge of the expected value distribution~\cite{INQ,Heterogeneous_Bitwidth_Binarization,outlier_aware}. However, none of these methods attempt to learn bitlengths leaving lots of potential untapped. 

One way of learning bitlengths is to use reinforcement learning (RL) to reduce the search space~\cite{Releq,HAQ}. However, after each new bitlength selection, RL algorithms need to estimate the accuracy to properly apply rewards. This step often requires a lengthy fine-tuning process for each selection. Consequently, this approach is not scalable (for finer granularities and larger networks) since the search space increases exponentially with the number of different datatype groups.
A better approach is to somehow map bitlenghts from the discreet into the continuous domain so that gradient descent can learn them. One successful approach is the recently proposed Mixed Precision DNN (MPDNN) that learns per layer bitlengths~\cite{Mixed_Precision_DNNs}. Given a hard memory constraint (total memory footprint needed to store all weights and the largest activation layer), MPDNN excels at finding the best way to utilize the allotted memory space  to maximize accuracy. However, MPDNN is not ideal for jointly optimizing bitlength and accuracy since without a memory constraint, MPDNN provides bitlengths that are significantly larger than necessary. In essence, to find an ideal bitlength distribution, the algorithm must be given in advance its resulting memory footprint.  Additionally, MPDNN does not allow for optimizing some other criteria such as operation count focusing solely on memory footprint. Our goal is precisely to fill these gaps. Ultimately, our method and MPDNN attempt to solve two very different problems: rather than best fitting within a memory capacity we minimize bitlengths and error concurrently  with benefits for compute and memory. 

We present \textit{BitPruning} a method that alleviates users from speculating bitlengths or memory footprints to jointly optimize bitlength and accuracy targeting both memory and compute benefits. By leveraging the power of training \textit{BitPruning} automatically \textit{learns} reduced bitlength (number of bits) integer representations at \textit{any} granularity (e.g., network, layer, or block of any shape or size). We demonstrate our extended training technique for integer representations as they have been shown to be sufficient for many DNN applications and because integer operations and functional units (multipliers and adders) are much less expensive area- and energy-wise than those for floating point. 

\begin{table}
\caption{Hardware accelerators that support variable bitlength activations ``A'' and weights ``W''.}
\label{tbl:Accelerators}
{
\resizebox{\columnwidth}{!}{
\begin{tabular}{l|lll}

\textbf{Accelerator}             & \textbf{Bitlength}           & \textbf{Target} & \textbf{Granularity}\\ \hline
Stripes~\cite{Stripes-MICRO} & Any & A&Layer\\
Dynamic Stripes~\cite{dynamicstripes} & Any & A&Group\\
Dpred~\cite{DPRed}& Any & A&Group\\
ShapeShifter~\cite{ShapeshifterMICRO}& Any & A&Group\\
BISMO~\cite{BISMO} & Any & W+A&Layer\\
Bit-Slicing~\cite{bilaniuk2019bit} & Any & W+A&Layer\\
BitFusion~\cite{Bit-Fusion} & Powers of 2 & W+A&Layer\\
Loom~\cite{LOOM}& Any & W+A &Layer\\
Bit Tactical~\cite{DBLP:conf/asplos/LascorzJSPMSNSM19}& Any & A &Layer/Group\\
Outlier-Aware~\cite{Park2018EnergyEfficientNN}& Any + outliers& W+A &Group\\
UNPU~\cite{Lee2018UNPUA5}& Any& W&Network\\
Proteus~\cite{Proteus}& Any& W+A&Group\\
GPUs~\cite{NVIDIA_TO}& 1, 4, 8& W+A&Group\\
CPUs w/ multimedia/vector& 1, 4, 8& W+A&Group\\

\end{tabular}
}}
\end{table}

\textit{BitPruning}'s goal is to squeeze out every possible benefit from reducing the bitlength used at \textit{any} desired \emph{granularity} and with \textit{any} \emph{priority}. Our method relies on an interpolation of bitlength to non-integer values allowing to
trade-off accuracy and bitlength by gradient descent, and it can be applied at any granularity: e.g., per network, layer, value, or any group of values in between, as long as it is statically determined. This allows us to target a variety of hardware platforms via a unified approach. 
Additionally, our approach allows groups to be emphasized arbitrarily to prioritize a selected criterion, such as multiply accumulate (MAC) operation count or memory footprint. This makes it possible to target different costs depending on the deployment scenario and workload, e.g., target computation for convolutional layers or memory for fully-connected ones. 
Finally, contrary to most approaches that ignore the first and last layers, ours is an \emph{end-to-end} method that optimizes the whole network from input to output.

Our contributions are that we develop 1)~an interpolation of integer bitlengths to non-integer ones, enabling bitlenghts to be learned, and 2)~a regularizer that can, during training, reduce the number of bits used whilst minimizing the effect on inference accuracy.

\section{Method}
\label{sec:method}

\textit{BitPruning} involves procedures for both the forward and backward passes used in training. We begin by defining a conventional, linear quantization scheme with integer bitlengths in the forward pass. This definition is then expanded to use non-linear bitlengths and we describe how this interpretation allows bitlengths to be learned using gradient descent. Subsequently, we introduce a parameterizable loss function, which enables \textit{BitPruning} to penalize larger bitlengths. Ultimately, we describe the final selection of integer bitlengths.

\subsection{Quantization}
The greatest challenge for learning bitlengths is that they represent discreet values over which there is no obvious differentiation. We overcome this fact by defining a quantization method based on non-integer bitlengths which is used during training. We start with a uniform integer quantization between the minimum and maximum of each layer and expand to non-integer bitlengths.

For an integer bitlength $n$ we use a  simple uniform fixed-point quantization between the minimum and maximum. That is, each float value $V$ is represented by an integer:

\[ Int(V,n)=Round((V-L_{min})/Scale(n))\]

where $Int(V,n)$ is the integer value with bitlength $n$, $L_{min}$ the minimum value in the layer (across the whole batch), and $Scale$ is the smallest representable difference:

\[Scale(n)=(L_{max}-L_{min})/(2^n-1)\] 
where $L_{max}$ is  the maximum value in the layer (batch). Consequently, this scheme quantizes an input float value $V$ to the following float value:

\[ Q_i(V,n)=L_{min}+Int(V,n)*Scale(n)\]

Throughout training, we represent the integer quantization as $Q(n)$. This quantization scheme does not allow the learning of bitlengths with gradient descent due to its discontinuous and non-differentiable nature. To expand the definition to real-valued $n$, the values used in inference during training are interpreted as an interpolation between the values represented by the nearest two integers:
 \[ Q_r(V,b+\alpha)=(1-\alpha) Q_i(V,b)+\alpha Q_i(V,b+1)\] 
where $n = b + \alpha$, with $0\leq \alpha < 1$, and $Q_i(V,b)$ is the integer bitlength quantization with $b$ bits.

The scheme can be, and in this work is, applied to activations and weights separately. Since the minimum bitlength per value is 1, $n$ is clipped at 1.0. This presents a reasonable extension of the meaning of bitlength in continuous space and allows for the loss to be differentiable with respect to bitlength. The final bitlength of each group for inference is then selected as the smallest integer greater or equal to the bitlength parameter learned during training.

During the forward pass the above formulae are applied to both activations and weights in the order shown. The values are converted to a floating point value that can be represented by an integer quantization, and if the bitlength is a non-integer value, the two nearest integer representations are interpolated. During the backward pass we use the straight-through estimator~\cite{bengio2013estimating,hubara2016binarized} to prevent propagating zero gradients that result from the discreteness of the rounding operation.

While training we assume that our model adequately represents the expressiveness of the network and has a monotonous relationship with accuracy. An exception to this assumption is observed when jumping from non-integer to integer bitlengths once the initial training phase is complete. This issue is resolved by extending the fine tuning phase, fixing the bitlengths and allowing accuracy to recover. Since this work targets inference, extending training is not considered a significant issue but an alternative solution may be the topic of future work.

\subsection{Loss Function}
The loss function penalizes bitlength by adding a weighted average (with weights $\lambda_i$) of the bits $n_i$ required for weights and activations of all layers. We define total loss $L$ as: \[ L=L_l+\gamma \sum(\lambda_i \times n_i) \] where $L_l$ is the original loss function, $\gamma$ is the regularization coefficient used for selecting how aggressive the quantization should be, $\lambda_i$ is the weight corresponding to the importance of the $i^{th}$ group of values, and $n_i$ is the bitlength of the activations or weights in that group. 

This loss function can be used to target any quantifiable criteria. For the majority of this paper, we select $\lambda_i$ such that the loss function weighs all layers equally and produces a loss of 1.0 for an 8-bit network. This is equivalent to stating that the benefit of reducing bitlengths is equal across all value groups. In Section~\ref{sec:experiments} we explore how $\lambda_i$ can be used to target a reduction of memory footprint of an activation and weight-heavy case, as well as minimize the number of MAC operations.

\subsection{Final Bitlength Selection}
The above training method will produce non-integer bitlengths which are meaningless for practical hardware. Hence, we adjust the learned non-integer bitlengths to the nearest greater integer. Tables~\ref{tbl:cifar10} and~\ref{tbl:imagenet} show that while this initially affects accuracy, continuing training recovers the accuracy loss. The final phase keeps bitlenghts constant and only updates the weights.

\section{Evaluation and Results}
\label{sec:experiments}

Without loss of generality we report experimental results for per-layer granularity, and for weights and activations separately. However, this choice is arbitrary and it is straightforward to adapt to any other granularity (e.g., per group of values that would be transferred or processed together). Similarly, it is a simple extension of the loss function to change the coefficients in the weighted sum to prioritize groups according to other quantifiable criteria.

\subsection{CIFAR10}

\begin{table*}[t!]
\centering
\caption{Activation/weight bitlengths and achieved accuracy of aggressive quantization and different strength reguralizers on CIFAR10 for non-integer and integer bitlengths.}
\label{tbl:cifar10}{
\begin{tabular}{l|l|lll|lll}
\multicolumn{2}{c|}{} & \multicolumn{3}{c|}{\textbf{Non-Integer Bitlengths}}& \multicolumn{3}{c}{\textbf{Rounded Integer Bitlengths}}  \\ 
\hline
\textbf{Network}             & \textbf{Regularizer}& \textbf{Accuracy} &\makecell{\textbf{Weights}\\ \textbf{\# of bits}}& \makecell{\textbf{Activations}\\ \textbf{\# of bits}}& \makecell{\textbf{Final}\\\textbf{Accuracy}} &\makecell{\textbf{Weights}\\ \textbf{\# of bits}}& \makecell{\textbf{Activation}s\\ \textbf{\# of bits}}\\ \hline
AlexNet &Baseline& \textbf{78.8}&32 float&32 float&\textbf{78.8}&32 float&32 float\\
 &$\gamma=$0.5& 78.3&~~3.78& ~~4.34&77.9&~~4.33&~~4.83\\
 &~~~~~~1.0&78.5&~~3.03&  ~~3.89&77.9&~~3.50&~~4.33\\\
 &~~~~~~2.5&78.3&~~2.45&~~3.18&75.0&~~3.00 &~~3.67\\
 &~~~~~~5.0&78.2&~~\textbf{2.06}&  ~~\textbf{2.72}&75.4&~~\textbf{2.50}&~~\textbf{3.17}\\
 &~~~~~~10.0&\multicolumn{3}{c|}{Does not converge}&\multicolumn{3}{c}{Does not converge}\\\hline
ResNet18 &Baseline& \textbf{94.9}&32 float&32 float&\textbf{94.9}&32 float&32 float\\
&$\gamma=$0.5& 94.2&~~1.67& ~~2.73&94.4&~~1.90&~~3.38\\
 &~~~~~~1.0&93.5&~~1.30&  ~~2.26&94.1&~~1.43&~~2.90\\\
 &~~~~~~2.5&93.1&~~1.15&~~2.01&93.4 &~~\textbf{1.24}&~~\textbf{2.43}\\
 &~~~~~~5.0&92.8&~~\textbf{1.14}&  ~~\textbf{1.99}&93.3&~~\textbf{1.24}&~~2.48\\
 &~~~~~~10.0&94.1&~~1.61&  ~~2.35&94.2&~~1.90&~~2.90\\\hline
\hline
\end{tabular}
}
\end{table*}

\begin{figure*}[t!] 
\centering
\subfloat[AlexNet]{\includegraphics[width=\half\linewidth]{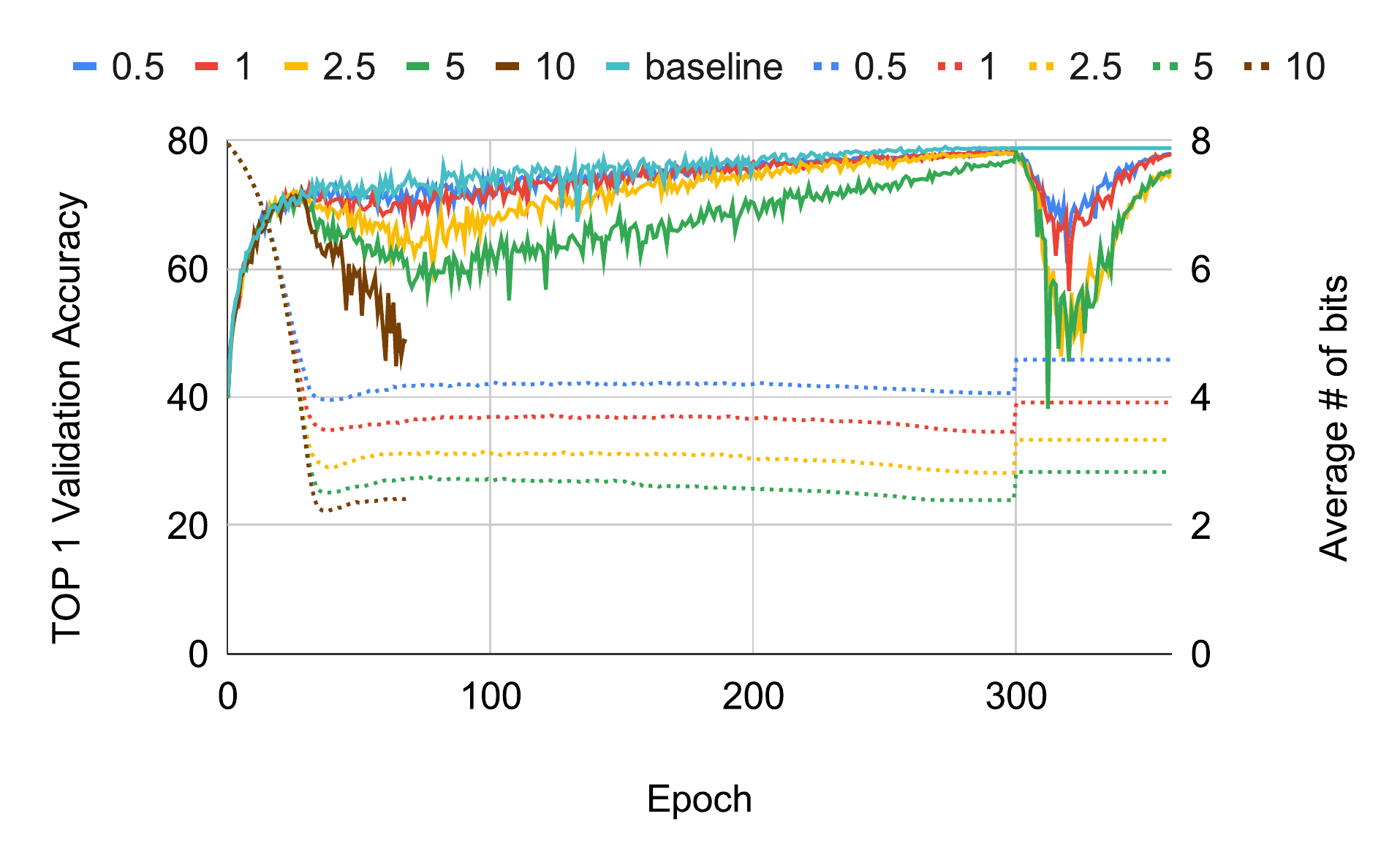}
\label{fig:alex1}
\label{fig:alex2}
}
\subfloat[ResNet18]{\includegraphics[width=\half\linewidth]{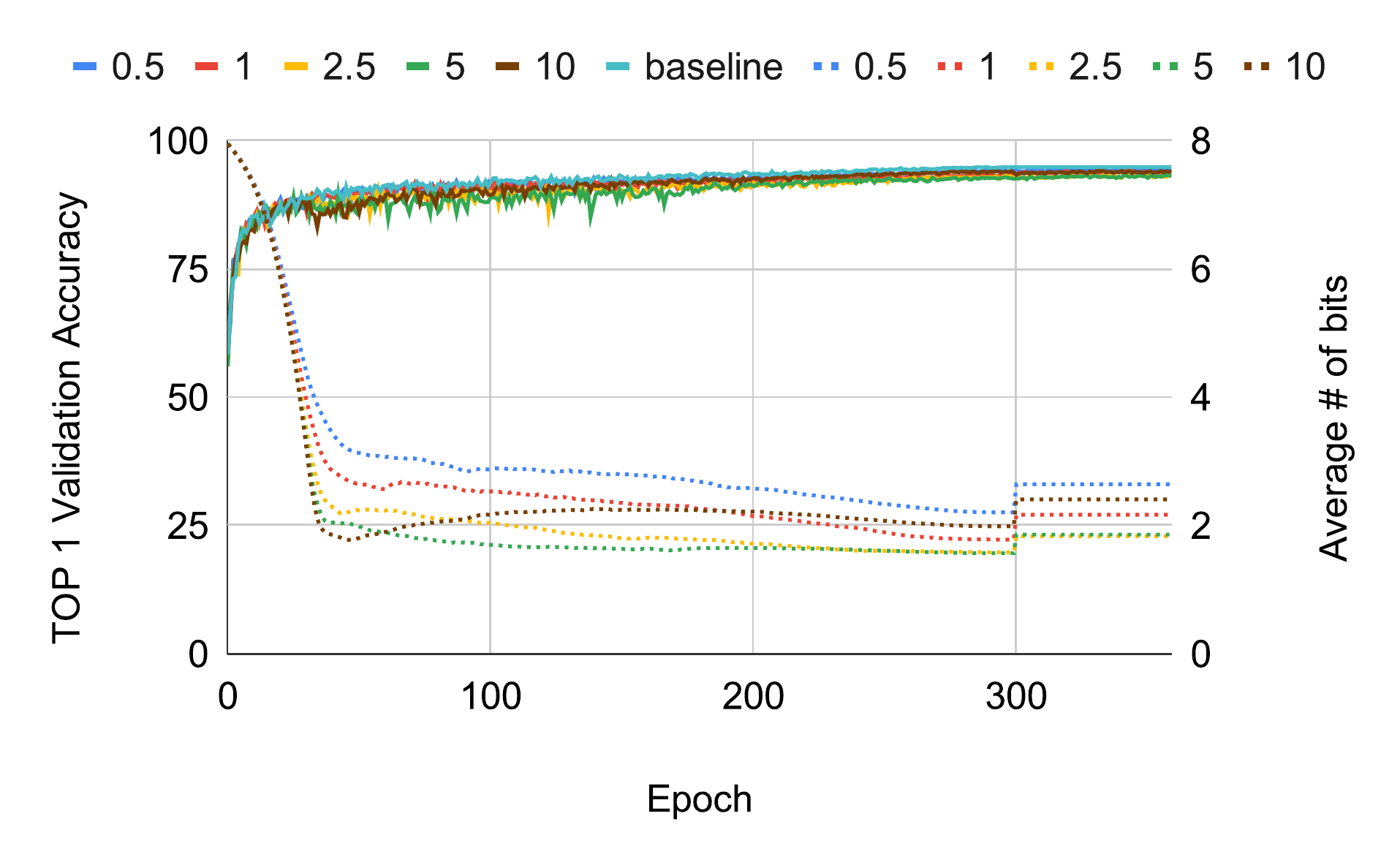}
\label{fig:res1}
\label{fig:res2}
}

\caption{CIFAR10 accuracy (solid) and bitlength (dotted) during training for different regularizers.}
\label{fig:cifar10}
\end{figure*}

This section discusses the BitPruning results for AlexNet~\cite{AlexNIPS2012} and ResNet18~\cite{he2016deep} on CIFAR10~\cite{cifar10}.

\subsubsection{Learning Bitlength}
The networks were trained over 300 epochs with the default fast.ai~\cite{howard2018fastai} parameters and one cycle policy in Pytorch~\cite{paszke2017automatic}. Table \ref{tbl:cifar10} shows TOP-1 validation accuracy for a 32b float baseline, as well as quantized models with our redefined loss. The bitlength weights ($\lambda_i$) are set to normalize all bitlengths to 8 bits and to emphasize all layers equally. If all layers use bitlength 8, loss will be $\gamma$. We change $\gamma$ to produce regularizers of progressively higher strength. Table~\ref{tbl:cifar10} reports the resulting average bitlegths over all layers.    Accuracies comparable to the baseline can be achieved with less than 3 bits on average for AlexNet, and 2 bits for ResNet. Progressively stronger regularizers achieve smaller bitlengths albeit at a slight degradation in accuracy. Weights consistently achieve smaller bitlengths than activations, while activations tend to benefit from more aggressive regularizers. This is expected as the weights are directly set by the quantization scheme, while activations are indirectly affected during run-time.

Figure~\ref{fig:cifar10} shows the validation accuracy and average bitlengths of activations and weights over the 300 training epochs. The bitlengths converge quickly and concurrently (all layers, weights and activations). While not shown in detail,  within 30-40 epochs the bitlengths across all groups drop near to their final values. Only slight changes during the remaining training epochs are observed. While all versions of ResNet18 closely track the baseline, AlexNet versions tend to follow closely the validation accuracy for first part of training. As bitlenths plateau, the accuracy drops for quantized networks with the degradation being greater for more aggressive quantizations. However, as training approaches the 300 epoch mark, all versions of AlexNet, except the most aggressive one, converge towards the baseline accuracy. For AlexNet more aggressive quantizers arrive to more aggressive bitlengths. The most aggressive attempt crashes at epoch 69. Similarly, all versions of ResNet18 reach average bitlengths corresponding to the aggressiveness of quantization, except for the most aggressive one. The most aggressive one dips the fastest, however it bounces off the minimum and converges to a bitlength that is larger than some of the less aggressive ones. This shows that the phase at which the Loss and the Regularizer conflict matters and affects both accuracy and bitlength.

Bitlengths vary noticeably per layer, creating more opportunities for specialized hardware; Uniform per network bitlengths would leave a lot of potential untapped. A finer quantization approach would presumably offer even more potential. Bitlengths show a slight descending trend towards latter layers.

\subsubsection{Selecting Bitlengths}
After this initial training, bitlengths are set to the ceiling of their learned value, resulting in a noticeable accuracy drop. There are two reasons why this occurs: our interpolation is imperfect (larger bitlengths often, but not always give better accuracy) and the network is partially trained for the smaller bitlength. Crucially, fine-tuning the networks regains this lost accuracy within 60 epochs. Table~\ref{tbl:cifar10} shows the drop in accuracy of integer bitlength networks as well as the effects of fine-tuning. In all cases, selecting the integer bitlengths increases the bitlengths by about 0.5 bits. Finally, Table~\ref{tbl:cifar10} shows the validation accuracy and average bitlengths of these fine-tuned integer bitlength versions in comparison with the baseline and non-integer bitlength versions. While the final integer bitlength ResNet18 versions slightly outperform the corresponding non-integer versions, the converse is true for AlexNet. Generaly integer and non-integer cases produce consistent accuracies.

\subsubsection{Other Architectures}
We demonstrate \textit{BitPruning} on a diverse set of architectures in Table~\ref{tbl:architectures_cifar10}, using the same training approach. It learns thrifty bitlenghts for all architectures with a good selection of hyper-parameters. 

\begin{table*}[t!]
\centering
\caption{Learning bitlenghts for different architectures}
\label{tbl:architectures_cifar10}{
\begin{tabular}{l|l|l|lll}
\textbf{Netwrok}& \makecell{Base\\ Accuracy} & \makecell{Quantized\\ Accuracy}&    \makecell{\textbf{Weights}\\ \textbf{\# of bits}}& \makecell{\textbf{Activations}\\ \textbf{\# of bits}}& \textbf{Regularizer}\\
     \hline
MobileNet V2~\cite{MobileNetV2}&94.6& 93.9 & 2.41& 3.86&0.5\\
DenseNet 121~\cite{DesnseNet}& 95.7& 94.4& 1.47& 1.97&1.0\\
DPN 92~\cite{DPN}& 95.7& 95.5&2.65&3.02&0.025\\
ResNext29(2x64d)~\cite{ResNext}& 95.3& 94.0& 1.11&2.11&1.0\\
PreAct ResNet18~\cite{PreActResNet}& 94.8&93.4 & 1.78& 2.64&1.0\\

\end{tabular}}
\end{table*}

\begin{table*}
\caption{Influence of loss function weighting on Cifar10. Average number of bits where the weights are selected as follows: BS - Batch Size, MAC - Multiply Accumulate}
\centering
\label{tbl:weighted}{
\begin{tabular}{l|ll|lll|lll}
\multicolumn{3}{c|}{} & \multicolumn{3}{c|}{Non-Integer bitlengths (Avg)}& \multicolumn{3}{c}{Rounded Integer bitlengths (Avg)}  \\ 
\hline
Network             & target & Accuracy &\makecell{BS of 1\\ footprint}& \makecell{BS of 128\\ footprint}& \makecell{MAC\\ operations}&\makecell{BS of 1\\ footprint}& \makecell{BS of 128\\ footprint}& \makecell{MAC\\ operations}\\ \hline
AlexNet&regular&78.5&2.97&3.22&3.82&3.40&3.64&4.25\\
&BS 1&\textbf{80.0}&\textbf{2.35}&3.53&5.03&\textbf{2.57}&3.8&5.45\\
&BS 128&78.7&2.42&\textbf{2.90}&3.91&2.80&\textbf{3.29}&4.33\\
&MAC ops&79.1&2.89&3.39&\textbf{3.39}&3.53&3.96&\textbf{3.90}\\

\hline
ResNet18&regular&93.5&1.23&2.85&2.26&1.35&2.70&2.15\\
&BS 1&\textbf{94.7}&\textbf{1.12}&2.85&1.78&\textbf{1.34}&2.70&2.16\\
&BS 128&93.6&1.67&\textbf{1.83}&2.55&1.86&\textbf{2.32}&3.10\\
&MAC ops&94.1&1.53&2.35&\textbf{1.74}&1.64&2.89&\textbf{2.07}\\
\hline
\end{tabular}
}
\end{table*}

\subsubsection{Network Structure}
\label{sct:Structure}

\begin{table*}[t!]
\centering
\caption{Number of bits for activations/weights and achieved accuracy for expanded/compressed layers. Bold numbers indicate change in int bitlength, Italic ones indicate change of non-int bitlength larger than 0.25.}
\label{tbl:structure_cifar10}{
\resizebox{\columnwidth}{!}{
\begin{tabular}{l|l|l
llllll|l|llllll|l}
&      & accuracy &      &      &      &      &      &      & avg  &      &      &      &      &      &      & avg  \\
      & base & 78.70    & 3.99 & 3.49 & 3.48 & 2.51 & 2.32 & 2.38 & 3.03 & 3.54 & 4.61 & 4.54 & 3.58 & 3.51 & 3.16 & 3.82 \\
     \hline
& 1.00 & 81.00    & \textit{3.17} & 3.33 & 3.42 & 2.52 & 2.26 & \textit{\textbf{3.22}} & 2.99 & 3.43 & 4.49 & 4.51 & 3.53 & 3.48 & \textit{\textbf{2.85}} & 3.72 \\
      & 2.00 & 79.60    & \textit{3.60} & 3.42 & \textit{\textbf{2.55}} & 2.40 & 2.18 & 2.17 & 2.72 & 3.46 & 4.54 & 4.40 & 3.50 & 3.44 & \textit{\textbf{2.65}} & 3.67 \\
x4    & 3.00 & 78.10    & 3.78 & 3.44 & 3.44 & 2.35 & 2.32 & 2.33 & 2.94 & 3.48 & 4.56 & 4.52 & 3.54 & 3.50 & 3.11 & 3.78 \\
      & 4.00 & 78.40    & 3.99 & 3.50 & 3.48 & 2.45 & 2.32 & 2.35 & 3.01 & 3.52 & 4.60 & 4.54 & 3.58 & 3.49 & 3.19 & 3.82 \\
      & 5.00 & 78.20    & 3.99 & 3.51 & 3.47 & 2.54 & 2.39 & 2.36 & 3.04 & 3.54 & 4.60 & 4.55 & 3.59 & 3.56 & 3.24 & 3.85 \\\hline
& 1.00 & 73.50    & \textbf{4.23} & \textit{\textbf{4.47}} & 3.50 & 2.44 & 2.47 & 2.44 & 3.26 & 3.62 & \textit{\textbf{5.51}} & 4.58 & 3.63 & \textbf{2.63} & \textit{3.40} & 3.90 \\
      & 2.00 & 77.20    & \textbf{4.09} & 3.54 & 3.54 & 2.52 & 2.42 & 2.44 & 3.09 & 3.57 & 4.63 & 4.61 & 3.61 & 3.53 & 3.36 & 3.89 \\
x0.25       & 3.00 & 79.10    & \textbf{4.06} & 3.52 & 3.45 & 2.54 & 2.39 & 2.49 & 3.07 & 3.56 & 4.64 & 4.55 & 3.59 & 3.54 & 3.16 & 3.84 \\
      & 4.00 & 78.50    & 3.95 & 3.49 & \textit{\textbf{2.60}} & 2.54 & 2.32 & 2.33 & 2.87 & 3.52 & 4.61 & 4.53 & 3.59 & 3.52 & \textbf{2.98} & 3.79 \\
      & 5.00 & 78.00    & 3.90 & 3.49 & 3.44 & 2.44 & \textit{\textbf{1.82}} & \textit{\textbf{1.54}} & 2.77 & 3.51 & 4.60 & 4.52 & 3.57 & 3.44 & \textbf{2.97} & 3.77
\end{tabular}}
}
\end{table*}

In this subsection we discuss the interplay of channel depth and bitlength. Prior work in ternary quantization has reported that increasing channel depth was essential in successfully training models to reduced datatypes~\cite{WRPN}. Presumably, deeper layers would lead to smaller bitlengths in the corresponding layers thus providing an additional knob for model designers to meet  execution time and energy constraints (e.g., doubling the number of channels while using 1/4 the bitlength reduces footprint to half). The experiments change the channel depth of each layer by x0.25 and x4 whilst all other layers are kept as-is and all networks are trained with the same parameters as the original one. Table~\ref{tbl:structure_cifar10} shows the validation accuracy as well as the bitlengths learned. On average the bitlengths are slightly smaller for the expanded networks and slightly larger for reduced ones. In some cases the change in bitlength  alters the final integer bitlegth. This change is usually, but not always, in the modified or the following layer. However, in most cases the differences are negligible compared to the cost of increasing the number of channels. The opposite can occasionally be observed with reducing channel depth. However, the accuracy can also vary significantly with these changes, and a set target could possibly be achieved with more aggressive regularization. Some of the benefits or costs of expanding/shrinking a layer appear to be absorbed by a increase or decrease of accuracy. As a result, it is not a straightforward task to balance channels/bitlengths whilst keeping accuracy constant. 

\subsubsection{Weighted Bit Loss}
Finally, we train AlexNet and ResNet18 with a modified loss function to minimize memory footprint or operation count. The bit loss is weighted respectively according to the number of elements or operations in each layer and for activations and weights separately. All other hyper-parameters are kept the same. We individually consider memory footprint for inference with batch sizes 1 and 128, representing a weight and activation-heavy network, operation count and original loss function weights. Table~\ref{tbl:weighted} shows the effects of the resulting loss functions for a range of criteria. The weighted bitlength regularizer allows us to successfuly target different criteria. On each criterion the targeted version of the network outperforms the generic case. 
However, differently weighted loss functions may affect accuracy and the final bitlength also depends on the final cutoff.

\subsection{ILSVRC2012}
The networks were trained on ImageNet~\cite{imagenet} over 180 epochs with default fast.ai~\cite{howard2018fastai} parameters and one cycle policy in Pytorch~\cite{paszke2017automatic}. Maximum learning rates were 0.01, 0.1 and 0.1 for AlexNet~\cite{AlexNIPS2012}, ResNet18~\cite{he2016deep} and MobileNet V2~\cite{MobileNetV2}. The 32 float baseline was trained over 90 epochs.

Table \ref{tbl:imagenet} shows TOP-1 validation accuracy for the 32-bit float baseline, and for quantized models with our loss. The loss parameters, $\lambda_i$ and $\gamma$, are set to weigh all layers equally and to normalize all bitlengths to 8. The table shows that accuracies comparable to the baseline for all networks can be achieved with around 3.5 bits on average for weights and around 4 bits for activations. Similarly to CIFAR10, the method is more capable to reduce bitlengths on the weights than activations.

\begin{figure*}[t!] 
\centering
\subfloat[AlexNet ]{\includegraphics[width=\third\linewidth]{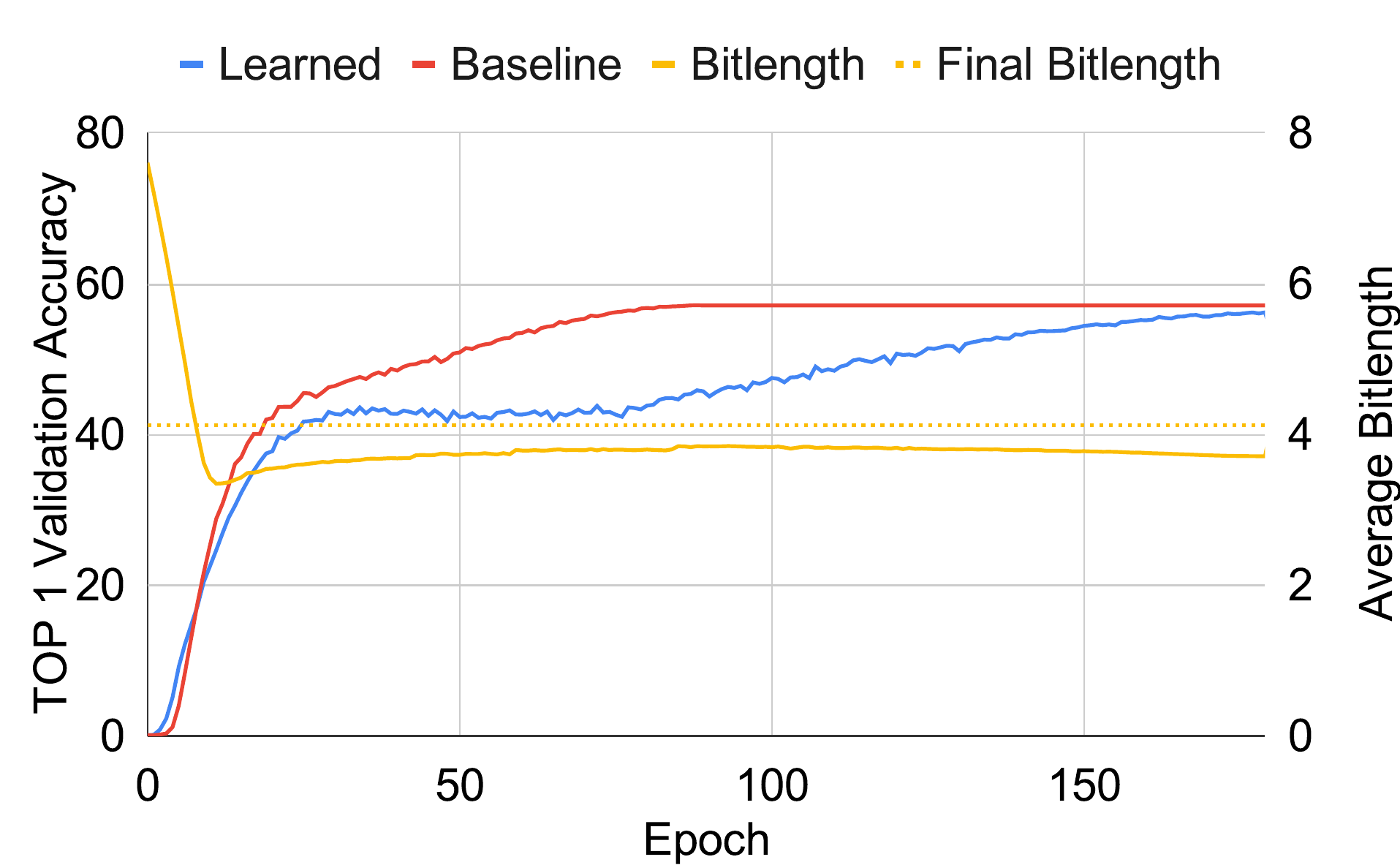}
\label{fig:image_alex1}
}
\subfloat[ResNet18]{\includegraphics[width=\third\linewidth]{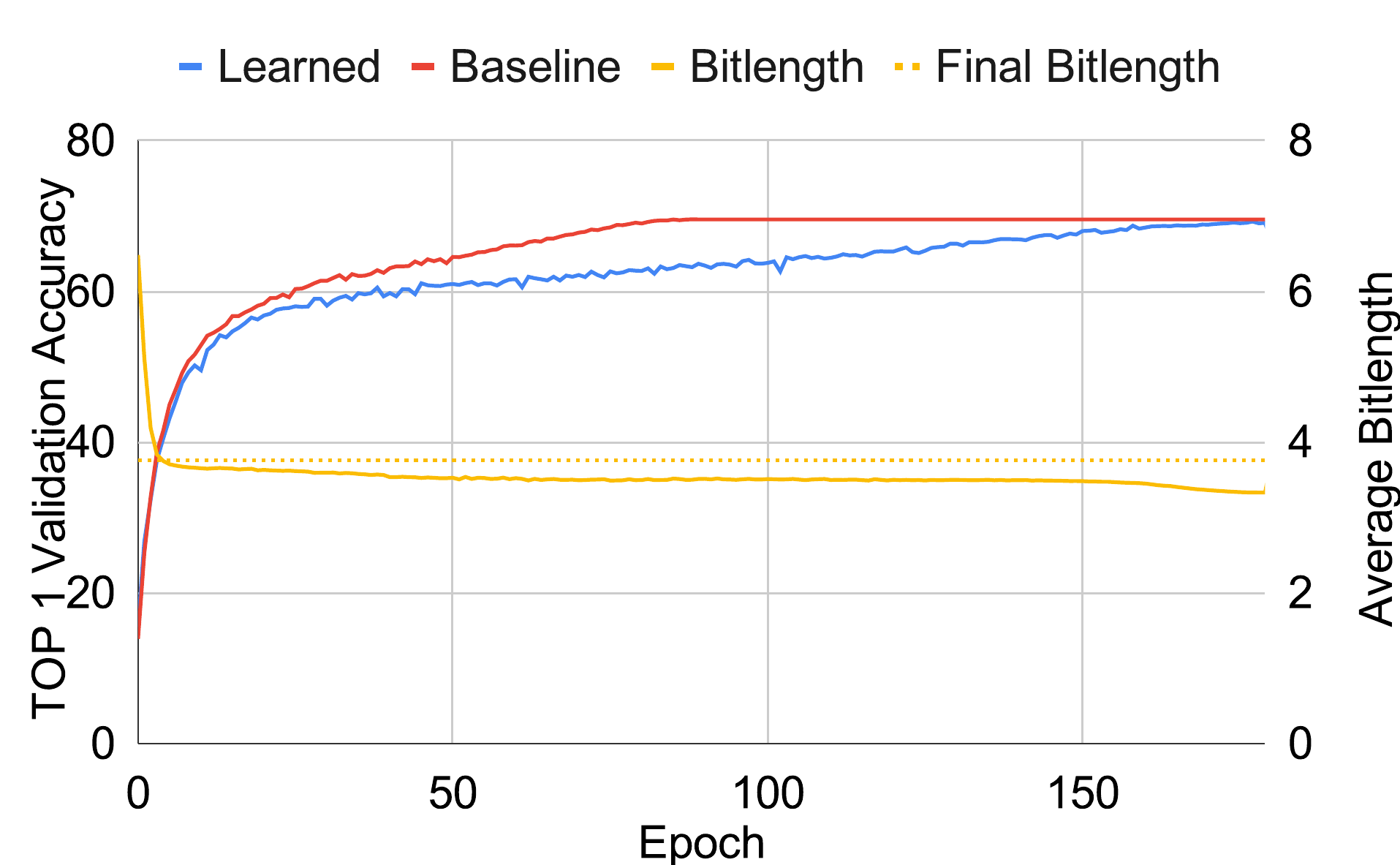}
\label{fig:image_res1}
}
\subfloat[MobileNetV2]{\includegraphics[width=\third\linewidth]{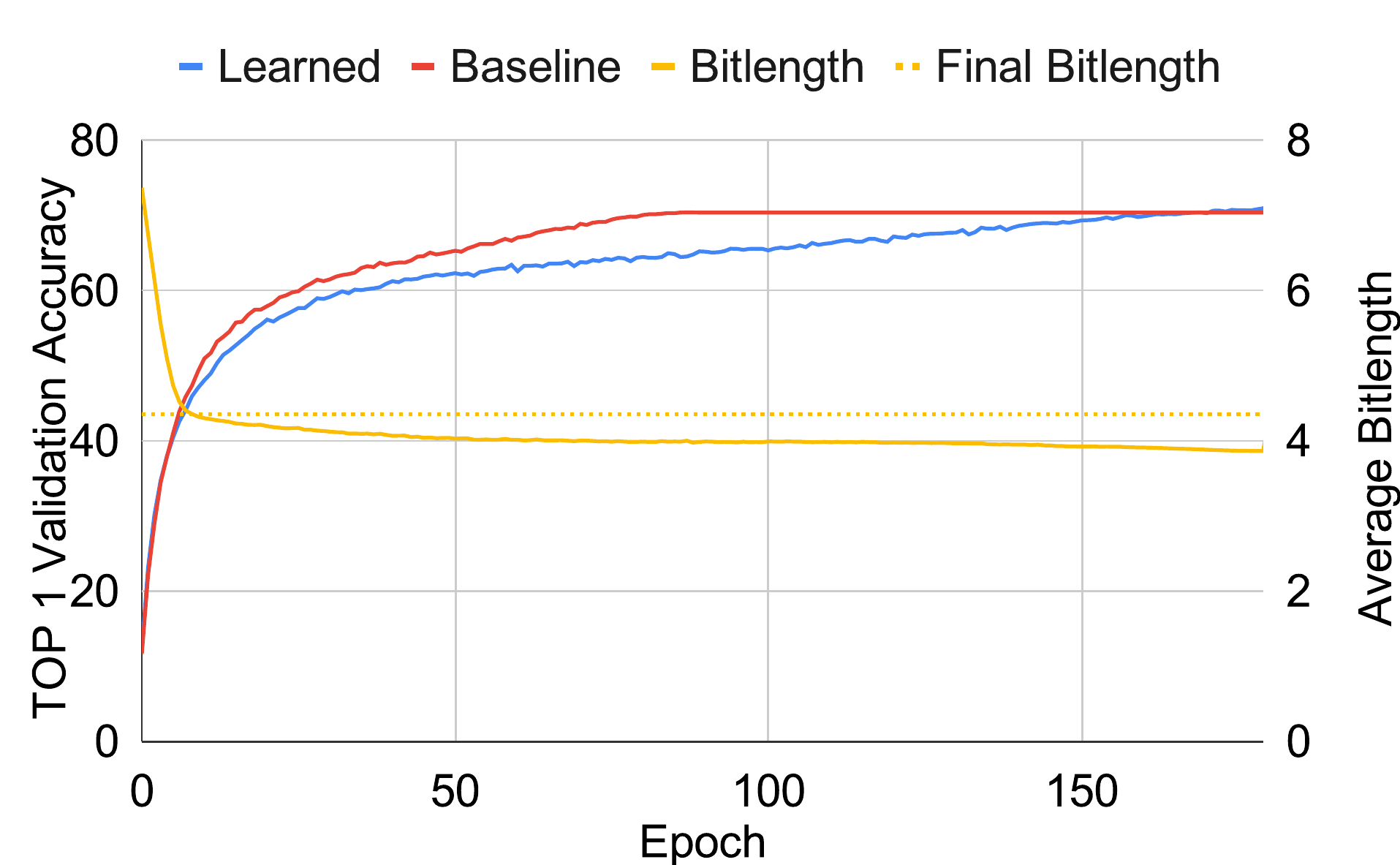}
\label{fig:image_mob1}
}

\caption{ImageNet validation accuracy and average bitlength during training.}
\label{fig:imagenet}
\end{figure*}

\begin{table*}
\caption{ImageNet bitlength and validation accuracy.}
\centering
\label{tbl:imagenet}{
\begin{tabular}{l|l|lll|llll}
\multicolumn{2}{c|}{} & \multicolumn{3}{c|}{Non-Integer bitlengths}& \multicolumn{3}{c}{Rounded Integer bitlengths}  \\ 
\hline
Network             & Regularizer &  Accuracy&\makecell{Weights\\ \# of bits}& \makecell{Activations\\ \# of bits}& \makecell{Final\\Accuracy}   &\makecell{Weights\\ \# of bits}& \makecell{Activations\\ \# of bits}\\ \hline
AlexNet&Baseline&\textbf{57.12}&32&32&\textbf{57.12}&32 float&32 float\\
AlexNet &1.0&56.20&3.35&4.07&55.07&3.88&4.38\\\hline
ResNet18 &Baseline&\textbf{69.54}&32&32&\textbf{69.54}&32 float&32 float\\
ResNet18 &1.0&69.26&2.86&3.80&69.19&3.38&4.14\\
\hline
MobileNet V2 &Baseline&70.44&32&32&\textbf{70.44}&32 float&32 float\\
MobileNet V2 &1.0&\textbf{70.99}&3.59&4.15&70.09&4.15&4.57\\
\hline
\end{tabular}
}
\end{table*}

\subsubsection{Learning Bitlengths}
Figure~\ref{fig:imagenet} shows the validation accuracy and average bitlengths of activations and weights over the 180 training epochs. After sufficient training the non-integer quantized networks reach near the baseline accuracy, however during training both quantized networks validation accuracy under-performs. During training per-layer bitlengths drop quickly and concurrently, within 10 epochs, to near final values. At this point the accuracy of AlexNets diverges, while the accuracy of ResNet starts diverging a bit later. While the bitlengths do not change much until the end of training, they slowly and noticeably increase for AlexNet, whilst for ResNet they continue to slowly decrease. Although the changes are small, they may tip the final integer bitlength.

\label{sec:imagenet_finalbitl}
\subsubsection{Selecting Bitlengths}
Final bitlengths are selected as the \textit{ceiling} of their learned values, and the network is finetuned for 90 (35 for MobileNet) epochs with 1/10\textsuperscript{th} of the learning rate. Table~\ref{tbl:imagenet} shows that this finetuning phase recovers validation accuracy loss due to selecting integer bitlengths.

\begin{figure*}[t!] 
\centering

\subfloat[AlexNet]{\includegraphics[width=\third\linewidth]{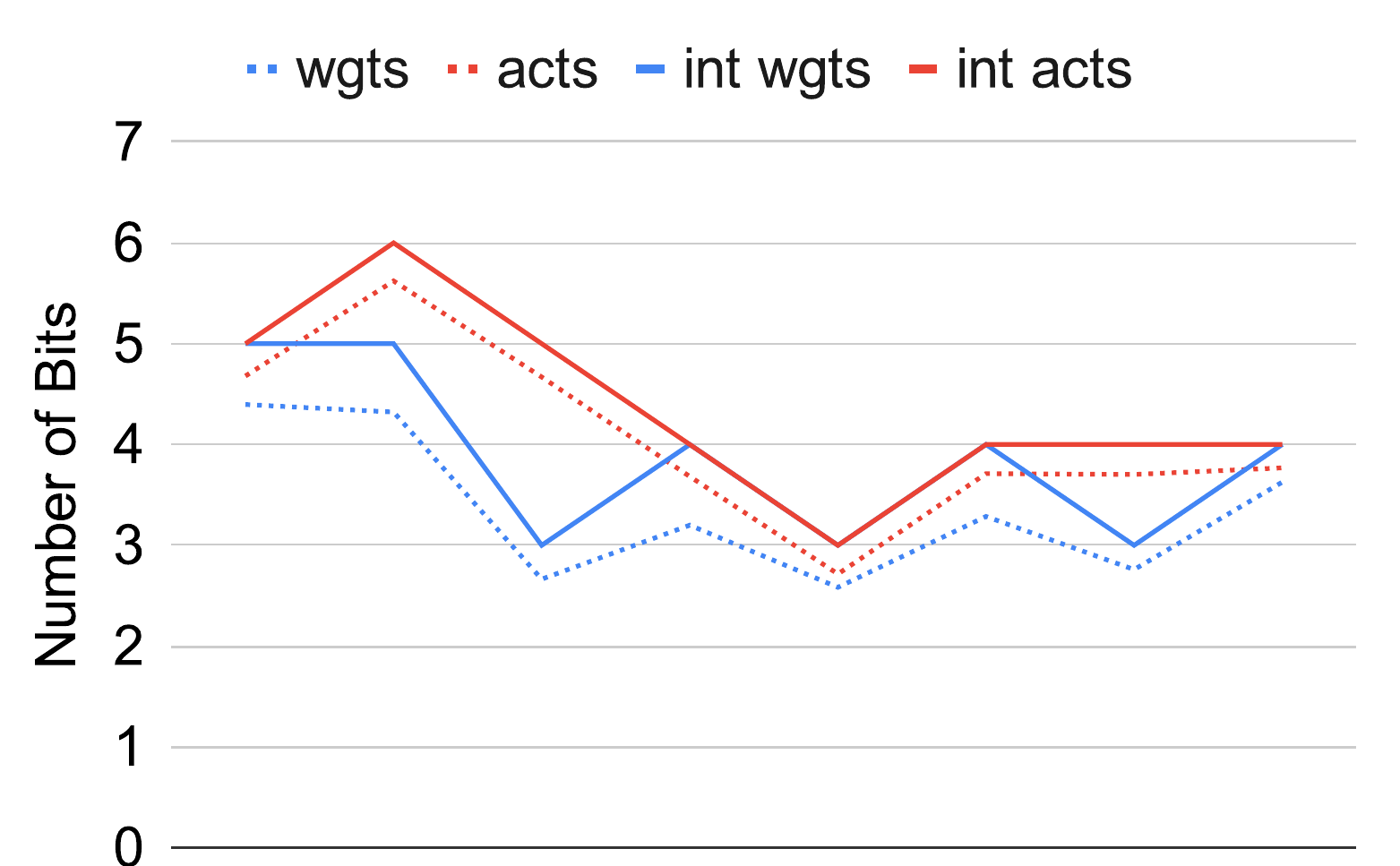}
\label{fig:alex1_pos}
}
\subfloat[ResNet18]{\includegraphics[width=\third\linewidth]{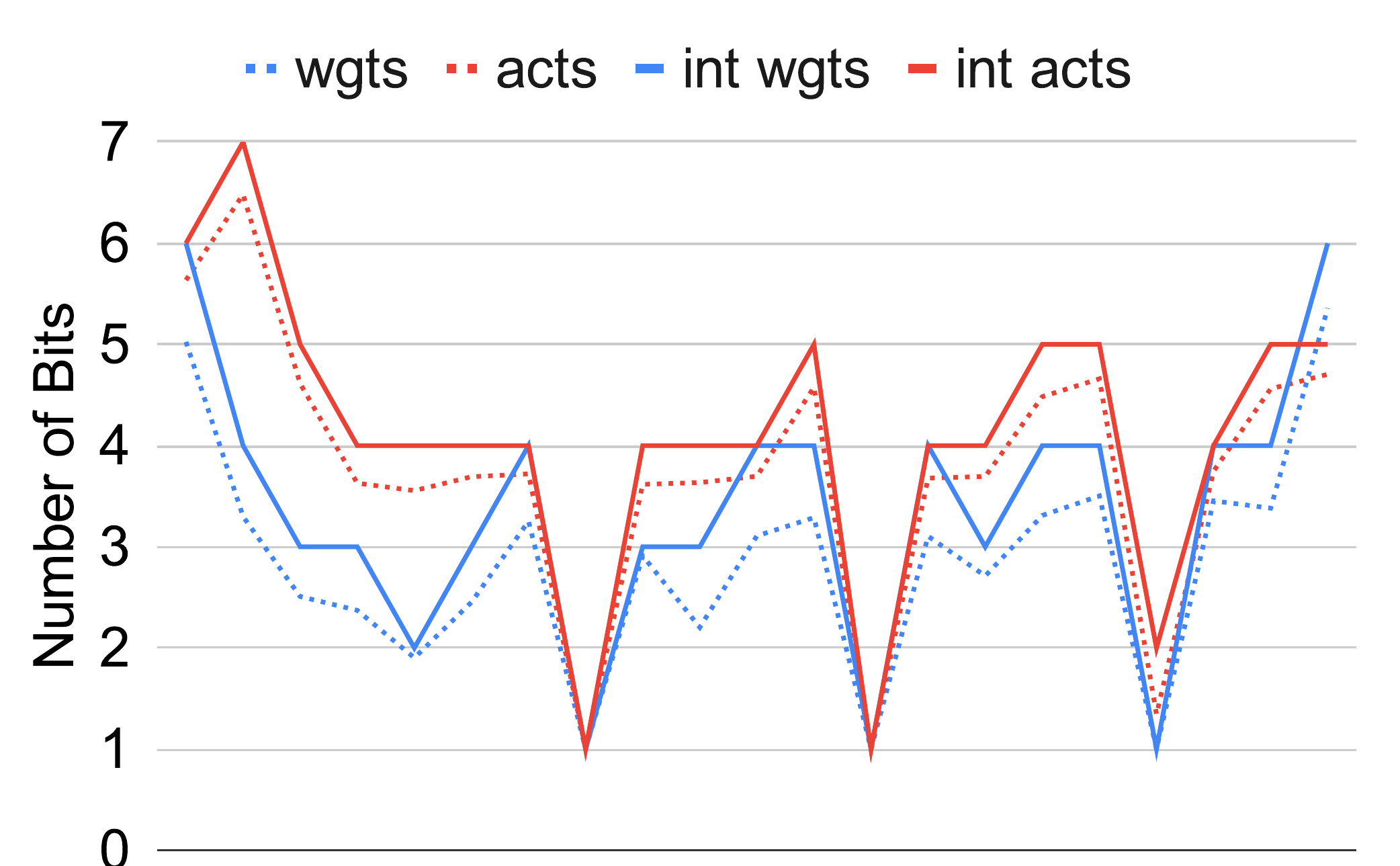}

\label{fig:res1_pos}
}
\subfloat[MobileNet V2]{\includegraphics[width=\third\linewidth]{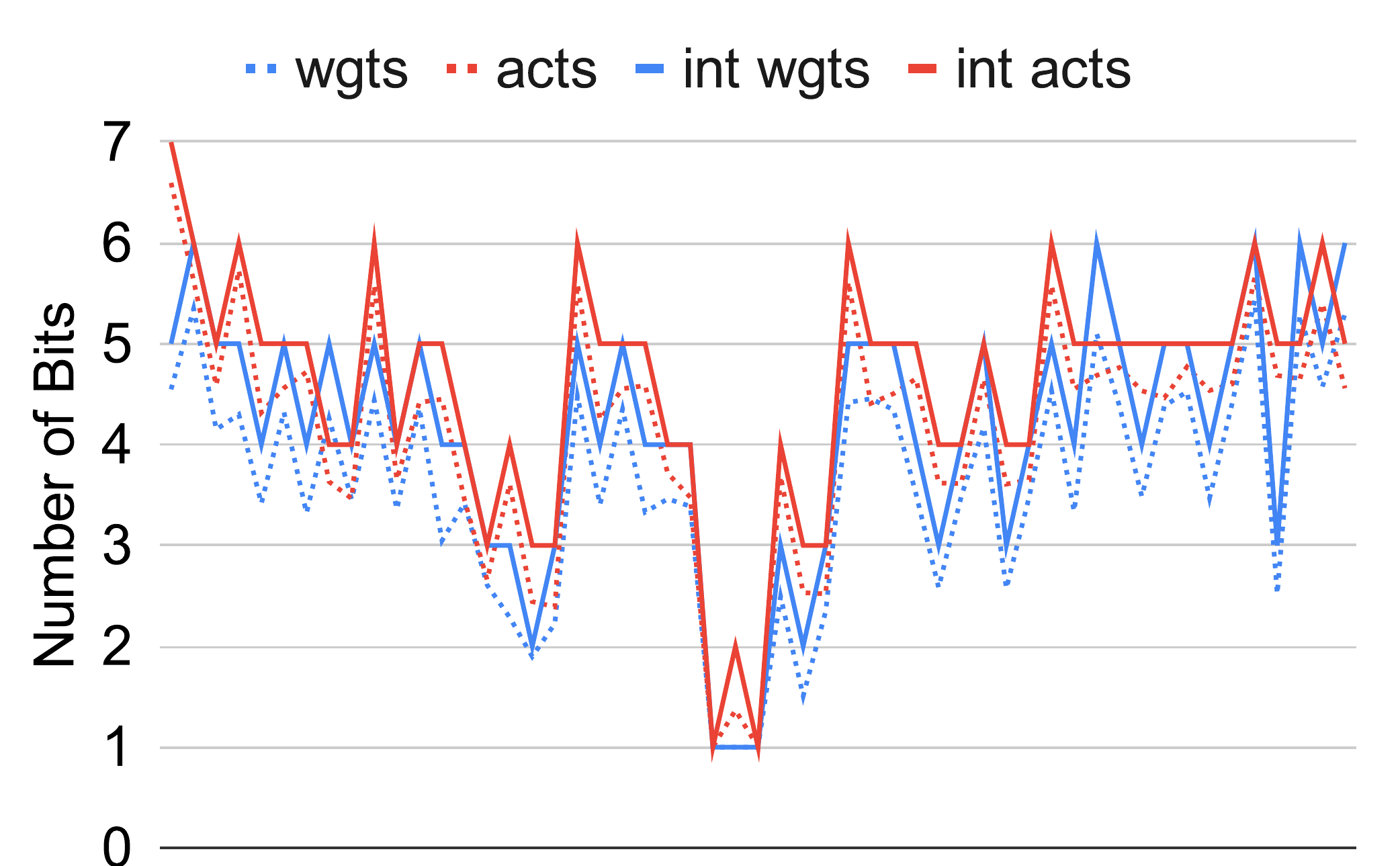}

\label{fig:mob1_pos}
}
\caption{Number of bits for activations/weights for each layer listed in a breadth first manner.}
\label{fig:imagenet_perlayer}
\end{figure*}

\subsubsection{Bitlength vs. Layer Position}
Figure~\ref{fig:imagenet_perlayer} shows that the bitlengths vary per layer for weights and activations, justifying the approach of using finer granularities. AlexNet and ResNet show a slight descending trend towards latter layers. Generally,  the first and last layers require more bits. Similarly, activations typically require larger bitlengths than weights.

\subsubsection{Selecting Bitlength Early}
We then explore when the final bitlengths can be selected. This is inspired by the fact the non-integer bitlengths converge quickly to near final values. Yet, the bitlengths slightly change throughout most of training. We test an early selection approach on AlexNet by training both the bitlengths and the model for their first 30 epochs, and then fixing bitlengths to integers over the next 150 epochs. This approach closely tracks the non-integer version with a final accuracy loss about 1\%. 

\subsubsection{Use as fine-tuning}

We also investigated using our method as means of fine-tuning pre-trained methods. 

We applied the bitlength training on an already trained 8-bit Alexnet with the same learning rate over 90 epochs. Just as in the original bitpruning version, the the drop of bitlengths significantly reduces accuracy of the fine-tuned one. After 90 epochs of training for bitlengths the networks starts to approach the baseline accuracy. 

However the bitlenghts of the fine-tuned version fall slower than the original. At the end of training the finetuned version requires 4.67 non-integer or 5.38 integer bits. This is significantly worse than the network trained with our regularizer from beginning. However, the results demonstrate that our method can be used to deliver benefits by fine-tuning pre-trained models.

\subsubsection{Comparison with Other Quantization Techniques}
Table~\ref{tbl:ComparisionWithOthers} clearly shows the advantage of our approach against uniform 4-bit (PACT) quantization and a profiled per-layer quantization in both validation accuracy and bitlength.

Additionally, we compare to MPDNN, a recent work which learns quantization parameters during training~\cite{Mixed_Precision_DNNs},  briefly discussed in Section~\ref{sct:introduction}. MPDNN starts with a pre-trained network and \textit{then} learns quantization scale and range to fit a given memory limit. Both techniques achieve a validation accuracy within 0.4\% of the full precision baseline for both ResNet and MobileNet V2 on ImageNet. When tasked with optimizing accuracy and memory footprint MPDNNs assigns 10.50MB/1.05MB  and 3.14MB/1.58MB of weight/activation memory for ResNet18 and MobileNet V2 respectively. As per the original MPDNN study, these memory requirements are for all weights and for the largest activation layer respectively. When MPDNN is tasked with optimizing accuracy while meeting a pre-determined memory constraint, the requirements drop to 5.4MB/0.38MB and 1.55MB/0.57MB. Note that these weight memory constraints must be expertly selected to ensure that near-baseline accuracy can still be achieved. With \textit{BitPruning} weights and activations respectively require 5.5MB/0.67MB and 2.2MB/0.72MB. \textit{BitPruning} achieves this low memory footprint despite: 1)~using a loss function that does not explicitly target memory footprint, and 2)~not requiring the user to specify a target memory budget that is known to work well. Further, Table~\ref{tbl:weighted} shows that by adjusting the loss function \textit{BitPruning} can explicitly target memory footprint. Finally, \textit{BitPruning} will better suit accelerators that benefit from reduced bitlengths, even after the network fits on chip.

\begin{table*}[!htbp]
\caption{Comparison with other quantization techniques}
\centering
\label{tbl:ComparisionWithOthers}{
\begin{tabular}{l|lll|lll|lll}
 & \multicolumn{3}{c|}{AlexNet}& \multicolumn{3}{c}{ResNet18}& \multicolumn{3}{c}{MobileNet V2}  \\ 
\hline
Method             & Accuracy &\makecell{Weights}& \makecell{Activations}& Accuracy &\makecell{Weights}& \makecell{Activations}& Accuracy &\makecell{Weights}& \makecell{Activations}\\ \hline
PACT&55.7&5.0&5.0&69.2&4.38&4.38&---&---&---\\
Profiled&55.78&7.63&5.75&65.56&6.41&6.34&69.9&7.33&7.02\\
BitPruning&55.07&3.875&4.375&69.19&3.38&4.14&70.09&4.15&4.57\\

\end{tabular}
}
\end{table*}

\subsubsection{Hardware Benefits}
Finally, in Table~\ref{tbl:Accelerators_effect} we show the benefits of our approach on existing and proposed hardware designs. \textit{BitPruning} significantly outperforms past profiling approaches \cite{milosispass19,judd:reduced}, as well as the 8 bit baseline.

\begin{table*}
\centering
\caption{Trained vs profiled quantization on select accelerators. Perf - Speedup, Mem - Total Storage}
\label{tbl:Accelerators_effect}
{
\begin{tabular}{l|ll|ll|ll|ll|ll|ll}
\hline
\multicolumn{1}{c|}{} & \multicolumn{4}{c|}{AlexNet}& \multicolumn{4}{c|}{ResNet18}& \multicolumn{4}{c}{MobileNet V2}\\
\multicolumn{1}{c|}{} & \multicolumn{2}{c|}{Trained}& \multicolumn{2}{c|}{Profiled}&\multicolumn{2}{c|}{Trained}&\multicolumn{2}{c|}{Profiled}&\multicolumn{2}{c|}{Trained}&\multicolumn{2}{c}{Profiled}\\\hline
Accelerator&Perf        & Mem        & Perf         & Mem        & Perf         &Mem         & Perf         &  Mem      & Perf         &Mem         & Perf         &  Mem      \\ \hline
Stripes&   $1.69\times$ & $0.95\times$ & $1.26\times$ & $0.98\times$ & $1.72\times$ & $0.94\times$ & $1.23\times$ & $0.98\times$ & $1.76\times$ & $0.53\times$ & $1.09\times$ & $0.95\times$ \\
Dpred&     $3.97\times$ & $0.91\times$ & $3.35\times$ & $0.91\times$ & $3.98\times$ & $0.89\times$ & $3.88\times$ & $0.90\times$ & $3.36\times$ & $0.35\times$ & $3.09\times$ & $0.38\times$ \\
BitFusion& $1.63\times$ & $0.60\times$ & $1.00\times$ & $1.00\times$ & $2.47\times$ & $0.53\times$ & $1.00\times$ & $1.00\times$ & $1.50\times$ & $0.67\times$ & $1.00\times$ & $1.00\times$ \\
Loom&      $3.74\times$ & $0.27\times$ & $2.81\times$ & $0.33\times$ & $4.11\times$ & $0.29\times$ & $3.44\times$ & $0.35\times$ & $3.70\times$ & $0.34\times$ & $2.67\times$ & $0.40\times$ \\
Proteus&   ---          & $0.47\times$ & ---          & $0.98\times$ & ---          & $0.50\times$ & ---          & $0.81\times$ & ---          & $0.53\times$ & ---          & $0.95\times$ \\

\hline
\end{tabular}
}
\end{table*}

\section{Training Costs}
\textit{BitPruning} incurs costs during training to learn the bitlengths. On ImageNet \textit{BitPruning} does not achieve baseline accuracy within the base 90 epochs ($1.5\%-2.5\%$  loss) and $2.3\times - 2.7\times$ more time, due to extra computations. To match the baseline accuracy, \textit{BitPruning} required 180 (2x) epochs and $4.6\times - 5.4\times$ more time. Additionally, \textit{BitPruning} requires twice as much memory. Neither of these are a major concern since \textit{BitPruning} targets aggressive quantization for efficient inference.

\section{Conclusion}
\textit{BitPruning} is capable of learning the bitlengths for accurate inference with controlled increase of loss function. On CIFAR10 we obtain 3.92 and 2.17 bits on average on AlexNet and ResNet18 respectively, with accuracies within 0.9\% for the weakest regularizer. On ImageNet we obtain 4.13, 3.76 and 4.36 average bitlengths with AlexNet, ResNet18 and MobileNet V2 respectively, whilst remaining within 2.0\%, 0.5\% an 0.5\% of baseline accuracy. \textit{BitPruning} can be applied at an arbitrary granularity with any selected weighted criteria. We demonstrate how this method is used effectively to minimize the compute workload and memory footprint in weight heavy (small batch) or activation heavy (large batch) cases. With CIFAR10 and for AlexNet and ResNet18 we reduce footprint by 10\% - 24\% and computation workload by 4\% - 8\%. Similarly, a modification of weights in the regularizer enables optimizing any other quantifiable criteria.
\textit{BitPruning} can be used to quantize all layers, including first and last, resulting in a simple end to end approach for quantization. Additionally, it naturally benefits existing hardware designs that can exploit different datatypes to significantly boost performance, and reduce memory traffic and footprint for all devices.

\section*{Broader Impact}

At the most fundamental level, all compute requires a hardware device that performs two functions: 1)~data storage and transfers (e.g., memory), and 2)~data manipulation/compute (e.g., addition, multiplication, etc.). Developing techniques that reduce energy  and execution time performance by targeting these fundamental operations is bound to impact virtually all segments of computing from the edge to the data center. The environmental costs are tremendous, and the potential benefits from more capable computing hardware are tremendous and have been reiterated time after time; without more powerful machines many of the innovations that we take for granted today would have never materialized and we are certain that many innovations to come will not materialize unless we success to keep improving computing hardware capabilities. The datatype used to represent and operate upon data can greatly impact silicon chip area, energy consumption, and as a result computing performance. It is for this reason, that this work, even though it targets a seemingly ``trivial'' and ``simple'' parameter, the datatype, can have an immediate, broad, and long lasting impact throughout. Moreover, choosing the right datatype is deceptively simple as the tradeoffs and potential usage scenarios (e.g., are we optimizing for a specific device, for a device to be built, for edge, for server, for memory on- and/or off-chip, for compute, etc.).

In more detail, the ability to learn the number of bits required for minimal uniform quantization but at a finer granularity than the whole network will optimize the execution time performance and energy costs of many commodity hardware platforms and specialized hardware accelerators that exploit this property. As a result the same work can be done with less power, decreasing the energy footprint of inference and reducing the climate impact of machine learning. Furthermore, the reduction of energy cost will enable to further push deep learning towards the edge, reducing the reliance on, and reducing communication with dataservers, and therefore improving user experience and  privacy as well as reducing energy cost on mobile devices.

The main drawback of the method is the extra required time and energy to train the model, which can be exacerbated for models that are continuously trained. These losses should be eclipsed by the benefits in most cases, however a more detailed analysis is need on a case by case basis.

\bibliography{references}
\bibliographystyle{IEEEtran.bst}
\end{document}